\documentclass[10pt,twocolumn,letterpaper]{article}

\usepackage{iccv}              

\definecolor{iccvblue}{rgb}{0.21,0.49,0.74}
\usepackage[pagebackref,breaklinks,colorlinks,allcolors=iccvblue]{hyperref}
\usepackage{arydshln}
\usepackage[accsupp]{axessibility}
\usepackage{subcaption} 
\usepackage{booktabs}
\usepackage{multirow}
\usepackage{pifont} 
\usepackage{bbding} 

\def\paperID{10215} 
\def\confName{ICCV}
\def\confYear{2025}

\title{Who is a Better Talker: Subjective and Objective Quality Assessment for AI-Generated Talking Heads}

\author{Yingjie Zhou\textsuperscript{\rm 1,2}~~~Jiezhang Cao\thanks{Corresponding authors.}\textsuperscript{\rm ~~3}~~~Zicheng Zhang\textsuperscript{\rm 1,2}~~~Farong Wen\textsuperscript{\rm 1,2} \\ Yanwei Jiang\textsuperscript{\rm 1,2}~~~Jun Jia\textsuperscript{\rm 1,2}~~~Xiaohong Liu\footnotemark[1] \textsuperscript{\rm 1,2}~~~Xiongkuo Min\footnotemark[1] \textsuperscript{\rm 1,2}~~~Guangtao  Zhai\footnotemark[1] \textsuperscript{\rm 1,2} \\
\textsuperscript{\rm 1} Shanghai Jiao Tong University 
\hspace{0.3cm} \textsuperscript{\rm 2} PengCheng Laboratory \hspace{0.3cm} 
\textsuperscript{\rm 3} Harvard Medical School
}

\begin{document}
\maketitle
\begin{abstract}
Speech-driven methods for portraits are figuratively known as ``Talkers" because of their capability to synthesize speaking mouth shapes and facial movements. Especially with the rapid development of the Text-to-Image (T2I) models, AI-Generated Talking Heads (AGTHs) have gradually become an emerging digital human media. However, challenges persist regarding the quality of these talkers and AGTHs they generate, and comprehensive studies addressing these issues remain limited. To address this gap, this paper \textbf{presents the largest AGTH quality assessment dataset THQA-10K} to date, which selects 12 prominent T2I models and 14 advanced talkers to generate AGTHs for 14 prompts. After excluding instances where AGTH generation is unsuccessful, the THQA-10K dataset contains 10,457 AGTHs. Then, volunteers are recruited to subjectively rate the AGTHs and give the corresponding distortion categories. In our analysis for subjective experimental results, we evaluate the performance of talkers in terms of generalizability and quality, and also expose the distortions of existing AGTHs. Finally, \textbf{an objective quality assessment method based on the first frame, Y-T slice and tone-lip consistency is proposed}. Experimental results show that this method can achieve state-of-the-art (SOTA) performance in AGTH quality assessment. The work is released at \href{https://github.com/zyj-2000/Talker}{\textcolor{magenta}{https://github.com/zyj-2000/Talker}}.

\end{abstract}
\begin{figure}[!t]
    \vspace{-0cm}
    \centering
    \includegraphics[width =1\linewidth]{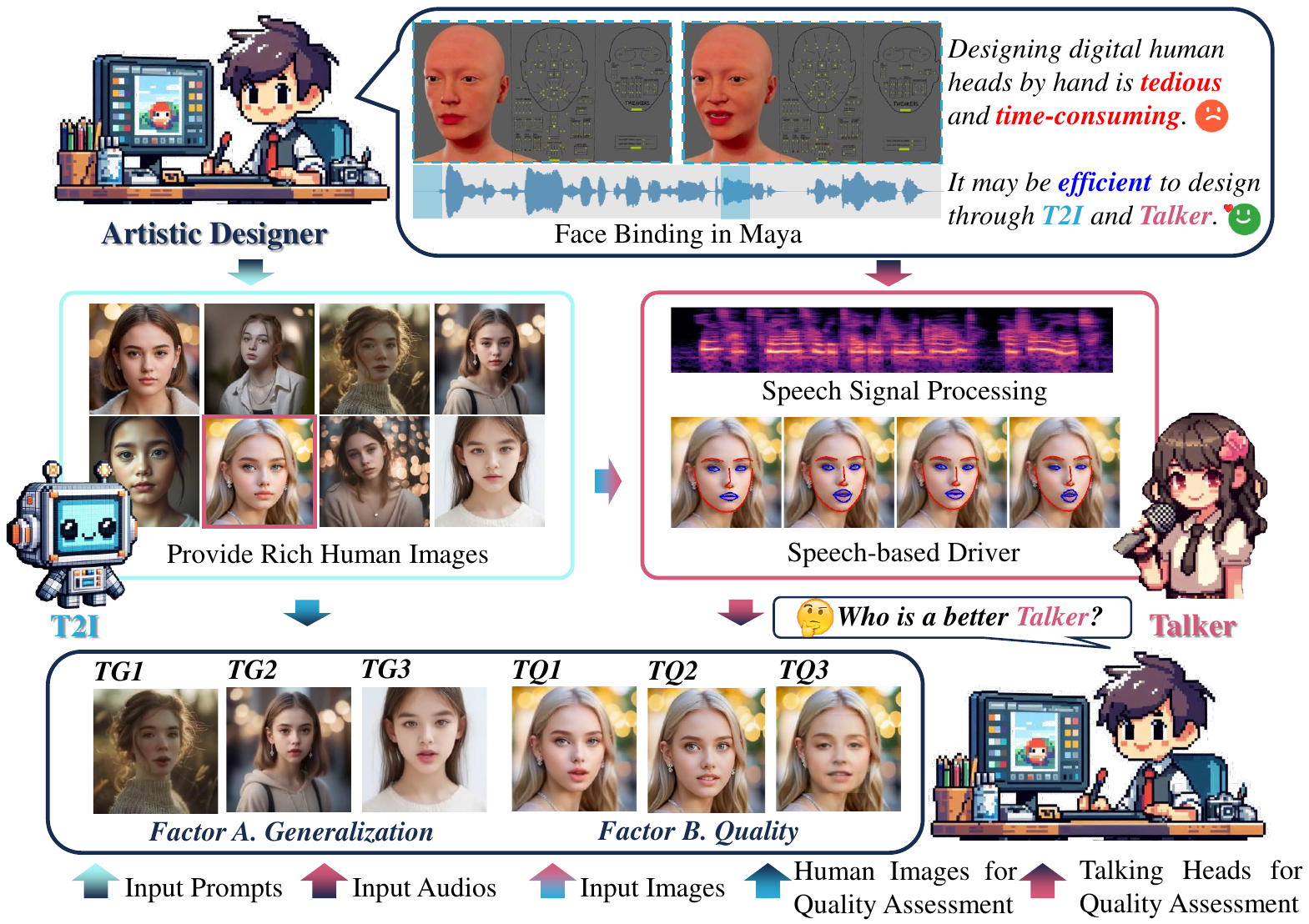}
    \vspace{-0.6cm}
    \caption{Manual approach to digital human head design versus Text-to-Image and Talker-based approaches.}
    \label{fig:teaser}
    \vspace{-0.7cm}
\end{figure}
\begin{table*}[!t]
\centering
\renewcommand\arraystretch{0.9}
\caption{The comparison of digital human databases and THQA-10K.}
\vspace{-0.4cm}
\resizebox{\linewidth}{!}{\begin{tabular}{cccccc}
\toprule
Database         & Modal   & Scale  & Methods &Distortions   & Description            \\
\midrule
DHH-QA \cite{dhhqa}    &Mesh + UV & 1,540 & Zhang et al. \cite{dhhqa}, Zhou et al. \cite{vitqa}& 7 model distortions  & Scanned Real Human Heads          \\
DDHQA \cite{ddhqa}   &Mesh + UV &800   & Zhang et al. \cite{zhang2023geometry}, Chen et al. \cite{chen2023no}&  7 model distortions and 2 motion distortions & Dynamic 3D Digital Human           \\
SJTU-H3D \cite{h3d}  &Mesh + UV &1,120  & Zhang et al.  \cite{h3d} & 7 model distortions   & Static 3D Digital Humans  \\ 
6G-DTQA \cite{6gqa}  &Mesh + UV& 400  & Zhang et al. \cite{6gqa} & 3 model distortions and 2 stream media issues   &   Dynamic 3D Digital Human  \\
THQA-3D \cite{thqa3d}&Mesh + UV& 1,000 & Zhou et al. \cite{thqa3d}& 5 stream media issues & Scanned Real Human Heads\\ 
CDHQA \cite{cdhqa}  &Video& 254  &None & 3 generative distortions  &Interactive Digital Human  \\ \hdashline
THQA \cite{thqa}  &Video + Audio& 800  & None & 9 generative distortions   &   AI-Generated Talking Heads  \\
AHQA \cite{imitator}  &Video& 1200  & Zhou et al. \cite{imitator} & 4 generative distortions   &   Animated Humans  \\

ReLI-QA \cite{reliqa}  & Image & 840 & None & 4 relighted methods & Relighted Human Heads \\
MEMO-Bench \cite{memo}  & Image & 7,145 & None & Sentimental Error & Emotional Human Heads \\
\hline
\bf{THQA-10K (Ours)} &\bf{Image + Audio}& \bf{10,457} &\bf{None} & \bf{10 generative distortions} & \bf{AI-Generated Talking Heads} \\
\bottomrule
\end{tabular}}
\vspace{-0.4cm}
\label{tab:databases}
\end{table*}
\section{Introduction}
\label{sec:intro}



Digital humans represent an emerging digital media technology focused on generating realistic representations of virtual characters endowed with human-like characters \cite{cumt}. Currently, the majority of high-quality digital humans are predominantly crafted and manipulated by skilled designers, necessitating extensive expertise and experience. In particular, the design process is cumbersome and time-consuming in terms of character modeling and facial animation. This method of manual design obviously suffers from low efficiency and high cost, hindering the popularization and promotion of digital humans. Fortunately, advancements in artificial intelligence (AI) have significantly facilitated the design of digital humans. On one hand, various types of text-to-image (T2I) models \cite{betker2023improving,midjourney,ideagram,flux,esser2024scaling,Rombach_2022_CVPR,podell2023sdxl,fooocus,kandinsky,opendalle,proteus} have been able to generate various types of character images, allowing for a more diverse appearance of digital humans. On the other hand, a variety of speech-driven methods \cite{makelttalk,audio2head,sadtalker,dreamtalk,wav2lip,videoretalking,dinet,iplap,musetalk,chen2024echomimic,wang2023seeing,yin2022styleheat,yang2024emogen}, which can be viewed as ``Talkers," have been developed to achieve the effect of Talking Head (TH). Although these talkers have improved the efficiency of digital human design, they inevitably face a variety of quality problems, which adversely impact the user experience. Therefore, it is imperative to evaluate these Talkers to provide objective and reliable reference metrics for various generative methods, thereby fostering the ongoing development of the digital human domain and enhancing user experience with AI-Generated Talking Head (AGTH) videos. More specifically, considering the AGTH generation process, two aspects should be considered to assess the quality of the talkers. \textit{\textbf{A) Generalization:}} A better talker should produce high-quality AGTHs across a wide range of portrait images. \textit{\textbf{B) Quality:}} A better talker must generate superior AGTHs for identical images and speech inputs.

Unfortunately, there has been limited research addressing these quality concerns and proposing credible solutions. To tackle this challenge, this paper first introduces the largest and most comprehensive AGTH Quality Assessment dataset named THQA-10K. This dataset encompasses a diverse selection of character materials, considering various ages and genders, and includes 14 prompts for character image generation. Each prompt is paired with five tailored speech sentences serving as driving audio. Furthermore, the dataset employs 12 leading T2I models and 14 talkers to generate AGTHs, resulting in a total of 10,457 instances within THQA-10K. Subsequently, volunteers are recruited to conduct subjective evaluations of the AGTHs, focusing on both distortion categories and visual quality scores. The findings reveal 10 distinct types of distortions among the AGTHs, alongside significant variations in quality depending on the talkers utilized. This highlights the critical need for quality assessments in this field. Leveraging the THQA-10K dataset and subjective ratings, we propose FSCD, an objective quality assessment method for AGTH based on the first frame, Y-T Slice \cite{shan2011xt} and tone-lip consistency. Experimental results validate the efficacy and superiority of this method, offering reliable objective metrics for the continued advancement of AGTH technologies. The principal contributions of this paper are as follows:

\begin{itemize} 
\item The THQA-10K dataset, comprising 10,457 AGTHs, has been constructed. To our knowledge, this dataset represents the largest collection created for the purpose of AGTH quality assessment, incorporating 12 prominent T2I models and 14 speech-driven methods. 
\item An objective quality assessment algorithm, referred to as FSCD, has been designed. This method integrates quality features from the first frame, Y-T slice, and tone-lip consistency to deliver an effective and robust objective evaluation of AGTH quality. 
\item The proposed THQA-10K dataset is representative and comprehensive to advance the field of digital human design. The designed FSCD method can achieve state-of-the-art (SOTA) objective quality assessment performance, which provides a reliable quality indicator for the field.

\end{itemize}

\begin{figure*}[t]
    \centering
    \vspace{-0cm}
    
    \subfloat[]{\includegraphics[width=0.345\linewidth]{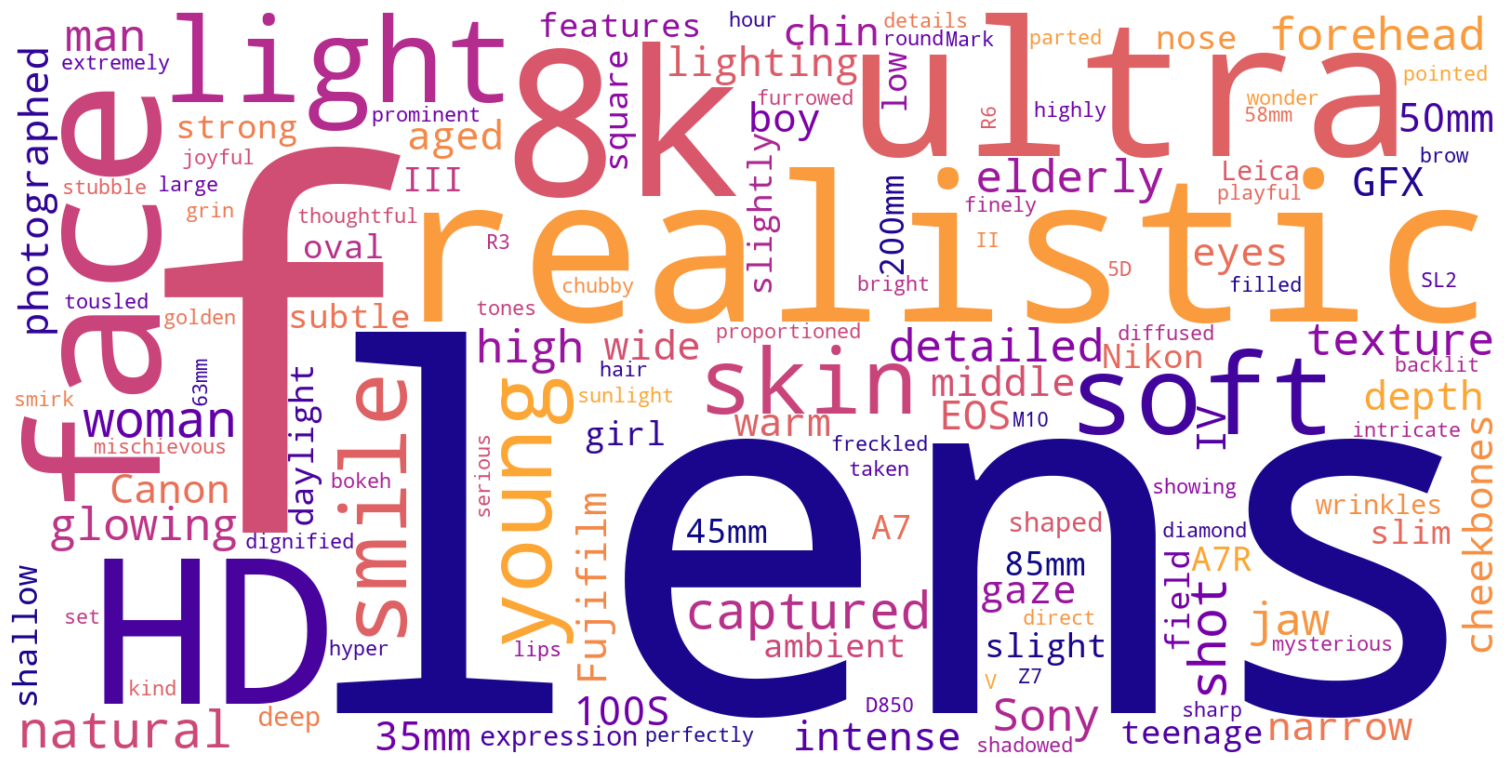}}
    \subfloat[]{\includegraphics[width=0.345\linewidth]{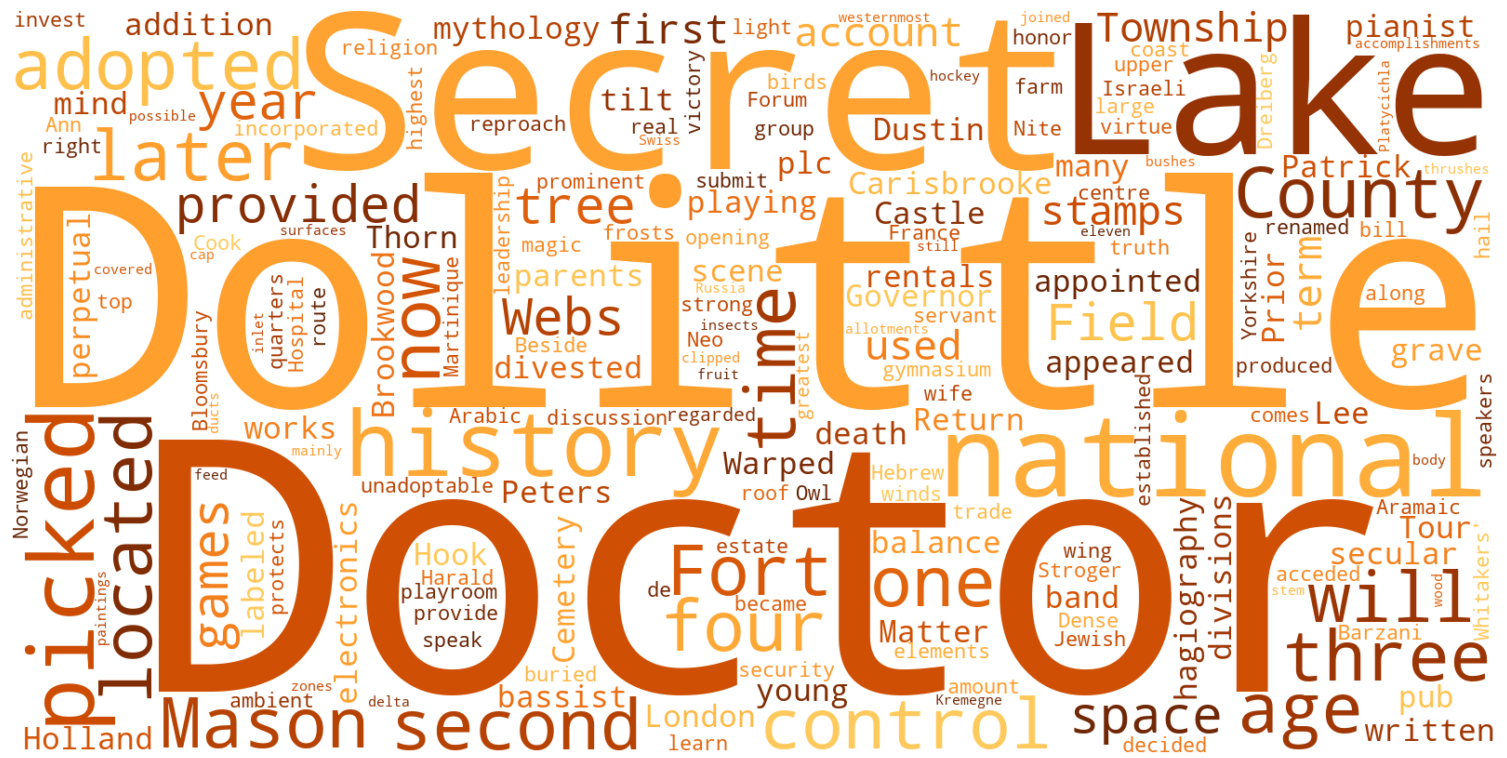}}
    \subfloat[]{\includegraphics[width=0.3\linewidth]{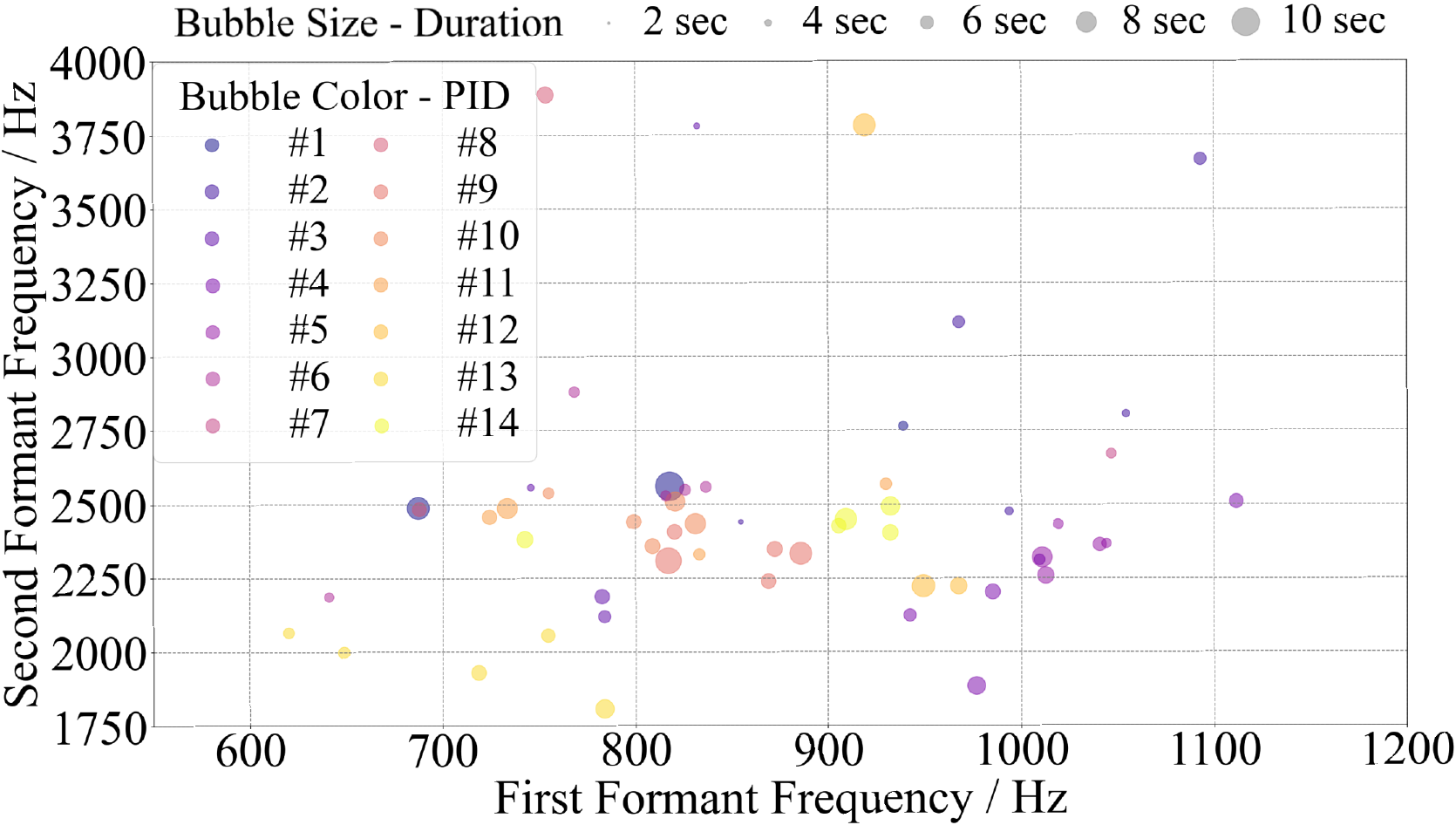}}
    \vspace{-0.4cm}
    \caption{Features of selected prompts and speeches. (a) Word cloud of selected prompts. (b) Word cloud of the speech text content. (c) Results of the resonance peak estimation for the selected speech. Speech samples that show resonance peak merging have been removed.}
    \label{fig:prompts}
    
    \vspace{-0.3cm}
\end{figure*}

\section{Related Works}
\label{sec:formatting}

\subsection{Talker: Talking Head Driven Methods}
Head-driven animation has been a significant area of research within the field of computer animation, primarily because the human head encompasses a wealth of facial details, expressive movements, and identity information. Traditional methods typically utilize computer-aided animation software, such as Maya\footnote{https://www.autodesk.com/products/maya/} or Blender\footnote{https://www.blender.org/}. These approaches require the binding of facial bones to the character model and the establishment of corresponding controllers to manipulate facial movements. Subsequently, keyframe animation techniques are employed, where appropriate keyframes are set for each controller to create various mouth shapes corresponding to different phonemes. This process is notably time-consuming and often necessitates extensive debugging. To address these challenges, the current standard solution in film and television production involves the use of facial capture sensors that track the key points of real THs \cite{ma2004realistic,fu2008real,schreer2008real,liu2025ntire}. The captured data is then imported into computer-aided animation software for further refinement and design. However, high-precision facial capture sensors are prohibitively expensive.

In recent years, advancements in AI have opened new avenues for the production of THs. The emergence of T2I models has not only simplified character image design but has also introduced speech-driven methodologies that directly apply speech to face images for AGTH generation. This approach alleviates a series of cumbersome processes and reduces equipment costs significantly. Speech-driven methods can be further classified into image-based  \cite{audio2head,sadtalker,audio2head,dreamtalk,eat,chen2024echomimic} and video-based \cite{wav2lip,videoretalking,dinet,iplap,musetalk,wang2023seeing,yin2022styleheat,yang2024emogen} techniques, depending on the input modality. Nonetheless, the absence of human design and oversight raises concerns regarding the quality of AGTHs, which is the central focus of this paper.

\subsection{Digital Human Quality Assessment}

With the development of digital human quality assessment of digital humans has become an emerging research component. As shown in Table~\ref{tab:databases}, many relevant datasets and objective quality assessment methods have been established, providing a rich data base and feasible solutions for the development of digital human quality assessment. Nonetheless, attention to AGTH is still lacking, with only Zhou $et$ $al.$ constructing a THQA dataset \cite{thqa}, which also fails to provide an effective quality assessment metric. Actually, PSNR and SSIM \cite{ssim} are still two commonly used quality metrics in the field of THs. However, it is clear that these two metrics are no longer suitable for the evaluation of AGTHs, due to the lack of corresponding reference videos. Although metrics such as Frechet Inception Distance (FID) \cite{fid}, LSE-C \cite{chung2017out}, LES-D \cite{chung2017out}, and CPBD \cite{cpbd} have also been used in the domain of THs, these metrics only focus on a certain dimension of AGTH and do not provide a comprehensive and effective assessment of AGTH. Therefore, there is an urgent need for effective objective indicators to measure the quality of AGTH in the field of AGTH. In Zhou $et$ $al.$'s experiments \cite{thqa}, they identified 9 common distortions present in AGTHs. However, the limitations of the THQA dataset are apparent. First, it consists of only 800 AGTHs, which constrains its ability to encompass AGTHs generated by a variety of contemporary models. Second, the exclusive reliance on StyleGAN \cite{stylegan,karras2020analyzing} for character image generation neglects the potential influence of current mainstream T2I models on quality. These constraints hinder the development of effective methods for objectively assessing AGTH quality. To address these deficiencies, this paper introduces a larger and more comprehensive THQA-10K dataset. Utilizing the THQA-10K dataset, we conduct extensive subjective experiments and design targeted objective quality assessment methods, thereby providing reliable reference metrics for AGTH quality evaluation.

\section{Database Construction}

\subsection{Prompts and Speeches}
To ensure diversity and representativeness among persons and to account for potential variations in AGTHs due to differences in gender and age, we select 14 distinct prompts for character portrait generation, each assigned a unique prompt ID (PID). Fig.~\ref{fig:prompts}(a) displays the frequency and content of these prompts, all of which are set to \textit{``8k"} quality to maximize the resolution and detail of the generated characters. Additionally, photographic elements are incorporated into some prompts to enhance realism. Overall, these prompts offer a comprehensive description of various human head features.

Aligned with the age and gender associated with each prompt, we select five corresponding speeches from the Common Voice speech dataset\footnote[3]{https://commonvoice.mozilla.org}. Each set of five speeches is sourced from the same speaker, ensuring consistent phonological features across the selected speeches for each prompt. To illustrate the diversity in speech content and features, we perform speech recognition and feature extraction on the selected audio samples, with results presented in Fig.~\ref{fig:prompts}(b-c). Several observations can be made from Fig.~\ref{fig:prompts}(b-c): 1) The speeches encompass a broad vocabulary, predominantly comprising common words, suggesting a rich variety of phonemes; 2) The first formant frequency of most speeches falls between 600 and 1150 Hz, indicating diverse mouth shapes during articulation. The second formant frequency ranges between 1750 and 4000 Hz, reflecting differences in tongue positioning during pronunciation; 3) For each PID, the five selected speech samples display clustering in their audio features, while the samples from different PIDs are more widely separated, indicating consistency within each PID; 4) Audio durations range from 3.06 to 10.12 seconds, providing a balanced representation of short interactions and longer conversations.

\subsection{Generative Models}
\begin{figure}[!t]
    
    \centering
    \includegraphics[width =1\linewidth]{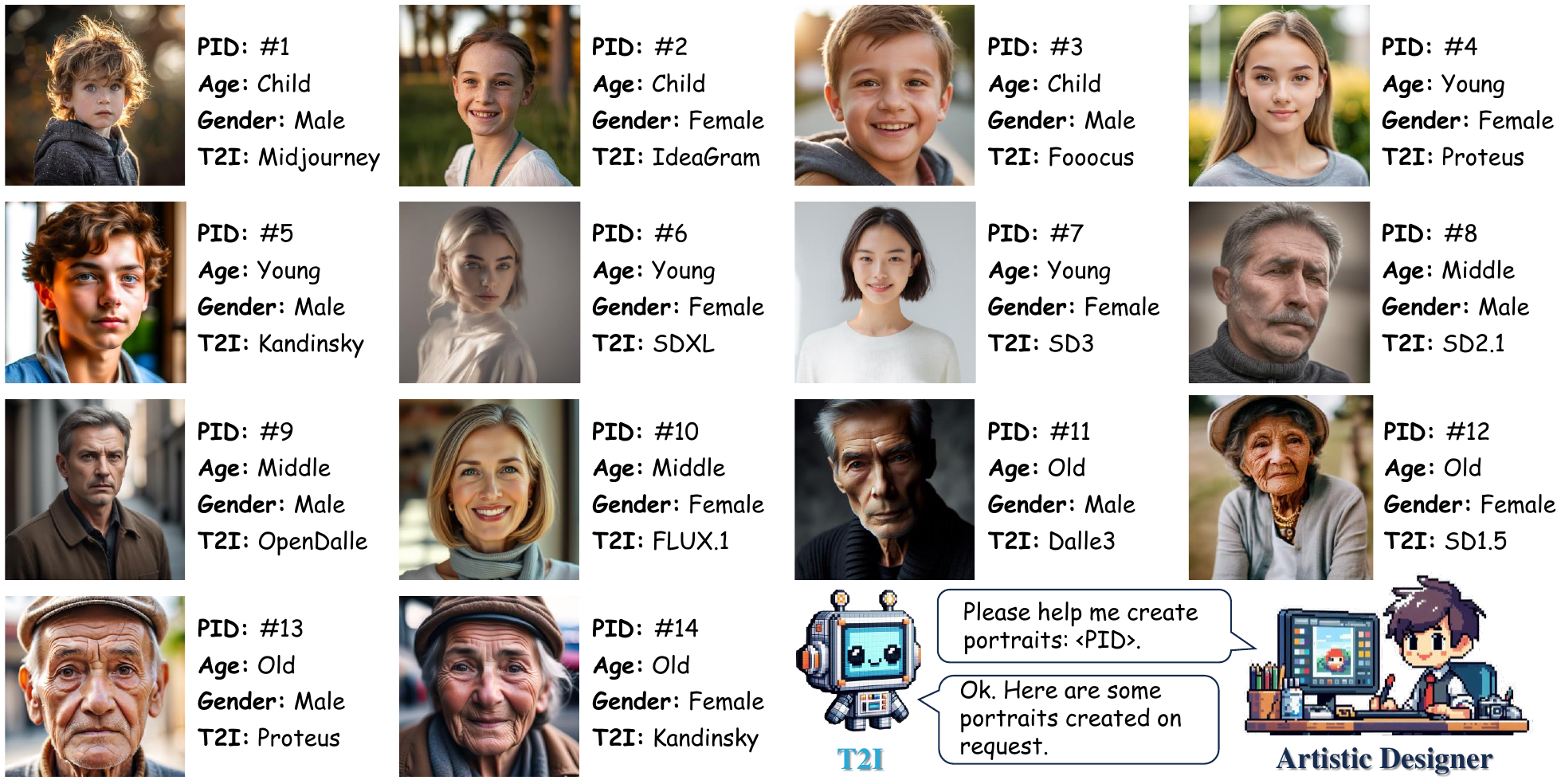}
    \vspace{-0.6cm}
    \caption{Generated portraits of different PIDs.}
    \label{fig:t2i}
    \vspace{-0.5cm}
\end{figure}

To generate AGTHs from prompts, two types of generative models are required. The first, text-to-image (T2I) models, are designed to produce portraits based on the provided prompts, serving as the foundational material for subsequent facial animation. Although previous research \cite{li2024aigiqa,zhang2024bench,aibench} has highlighted variations in the quality of images produced by different T2I models, there has been limited targeted discussion addressing the quality of generated portraits. To comprehensively evaluate the generalization performance of talkers and to investigate the impact of differences in the quality of generated portraits on AGTHs, we select 12 prominent T2I models for image generation. Details regarding the selected T2I models are presented in Table~\ref{tab:t2i}, with a subset of the generated portraits illustrated in Fig.~\ref{fig:t2i}.


\begin{table}[!tp]
    \centering
    \caption{Details of T2I models employed for generation. }
    \vspace{-0.3cm}
    \resizebox{1\linewidth}{!}{\begin{tabular}{c|c|c|c|c}
    \toprule
         Type & Label &T2I Model & Year & Output Resolution\\ \hline
          \multirow{3}{*}{Closed source} &DL3 &Dalle3 \cite{betker2023improving} &2023  & 1,024×1,024\\   
          &MJ6&MidjourneyV6 \cite{midjourney}&2023  & 1,024×1,024\\ 
          &IDG&Ideagram \cite{ideagram}&2024  & 1,024×1,024\\  \hdashline
          \multirow{9}{*}{Open source} &SD2&Stable Diffusion 2.1 \cite{Rombach_2022_CVPR}&2022  & 512×512\\
          &SD1&Stable Diffusion 1.5 \cite{Rombach_2022_CVPR}&2022  &  512×512\\
          &SDX&Stable Diffusion XL \cite{podell2023sdxl}&2023  &  1,024×1,024\\
         &FCS&Fooocus \cite{fooocus}&2023  & 1,024×1,024\\
         &KDS&Kandinsky \cite{kandinsky}&2023  & 1,024×1,024\\
         &ODE&OpenDalleV1.1 \cite{opendalle}&2023  & 1,024×1,024\\
         &PTS&Proteus \cite{proteus}&2024  & 1,024×1,024\\ 
          &FLU&FLUX.1 \cite{flux}&2024  & 1,024×1,024\\
          &SD3&Stable Diffusion 3 \cite{esser2024scaling}&2024  & 1,024×1,024\\

    \bottomrule
    \end{tabular}}
    \label{tab:t2i}
    \vspace{-0.3cm}
\end{table}

Another critical generative model is represented by speech-driven methods, commonly referred as talkers. To ensure that the constructed dataset encompasses a wide range of existing mainstream driving methods, 14 talkers are utilized to target the generated character images for AGTH generation. Table~\ref{tab:talker} provides details of the selected talkers, including their respective output resolutions. It is important to note that variations in output resolution exist and the selected driving methods are totally implemented using source code provided by the original authors. Additionally, for the video-based driving methods, the input video consists of a repeated driving image with a duration set to one second and a frame rate of 25 frames per second. Ultimately, a total of 10,457 AGTHs are successfully generated for 14 prompts and their corresponding 70 speeches, culminating in the creation of the THQA-10K dataset.

\subsection{Data Statistics and Subjective Experiment}
\begin{figure}[!t]
    \centering

    \subfloat[]{\includegraphics[width=0.333\linewidth]{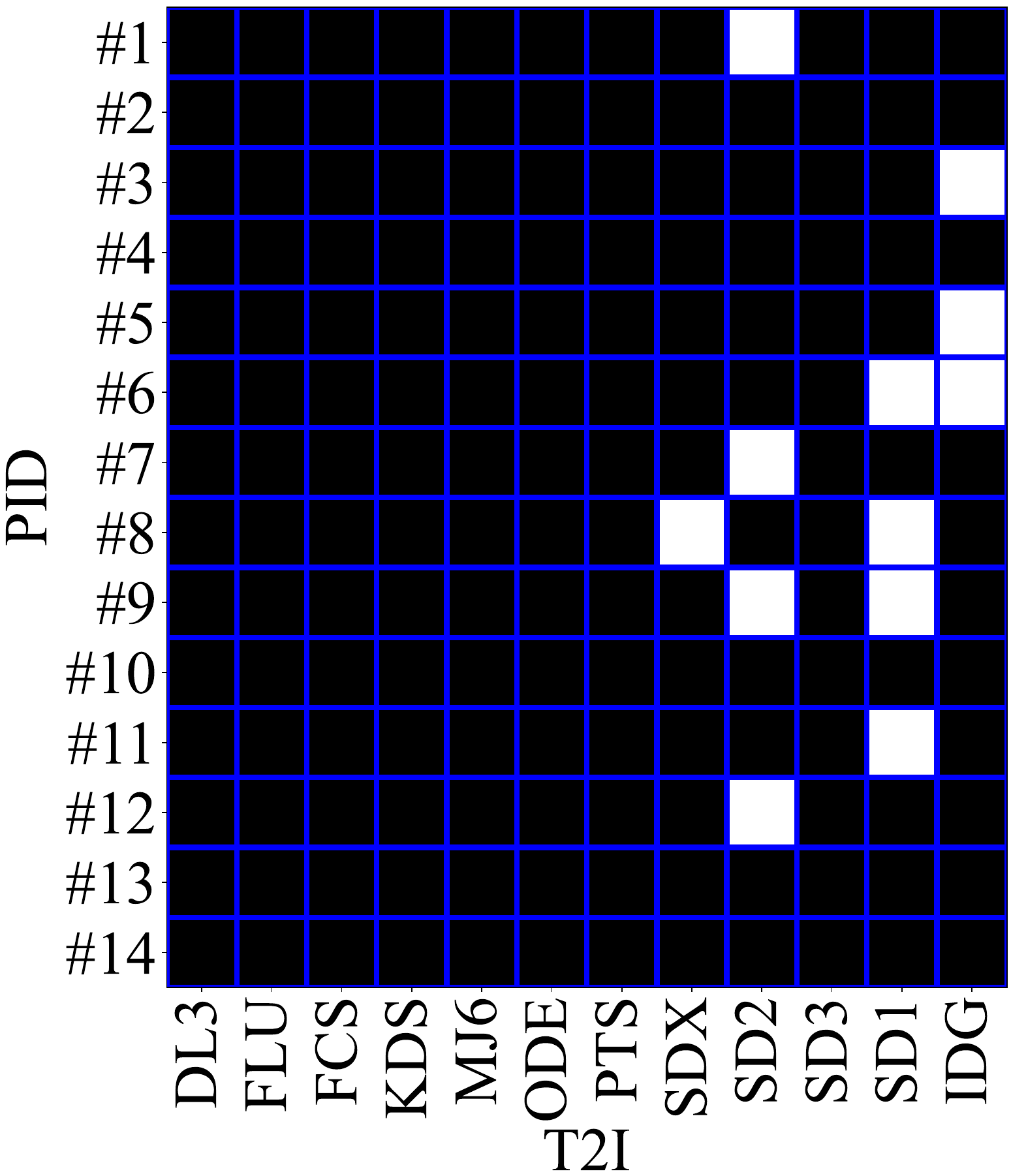}}
    \subfloat[]{\includegraphics[width=0.666\linewidth]{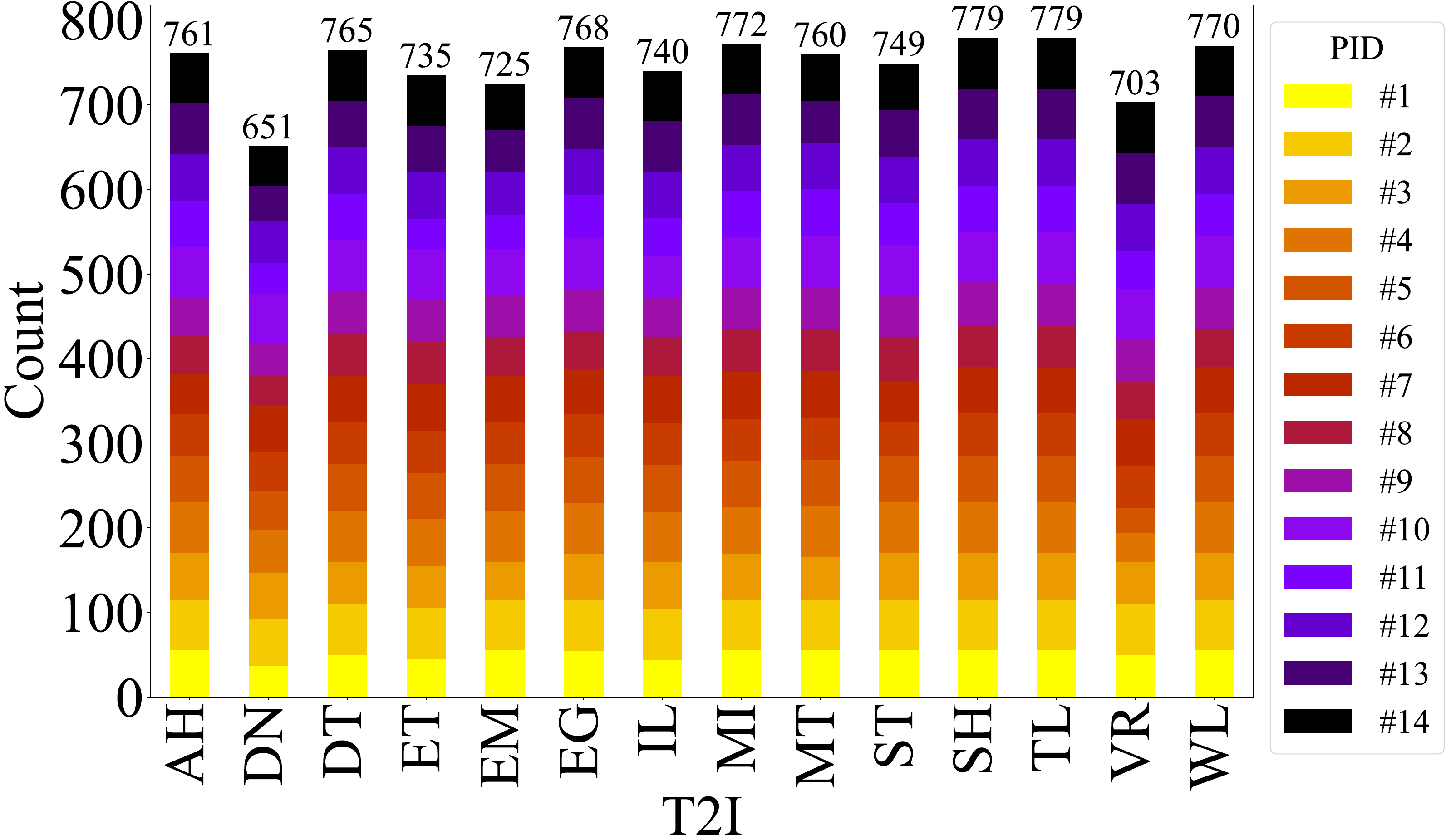}}
    \vspace{-0.2cm}
    \caption{Visualization of the number of successful generation. (a) Portraits that can be successfully generated. The black block indicates success while the white one indicates failure. (b) Number of AGTHs that can be successfully generated.}
    \label{fig:success}
    \vspace{-0.5cm}
\end{figure}

To assess the generalization capability of various T2I models and talkers, we conduct a statistical analysis of the number of portraits and AGTHs successfully generated by these generative models, as illustrated in Fig.~\ref{fig:success}. From Fig.~\ref{fig:success}(a), it is evident that the two T2I models, SD1 and SD2, demonstrate limited effectiveness in generating portraits across a range of prompts. However, for IDG, its' suboptimal performance can be attributed to the constraints posed by prompt-sensitive vocabulary. Additionally, Fig.~\ref{fig:success}(b) highlights the generalization performance of different talkers. Notably, the SH and TL talkers exhibit the strongest generalization abilities. Remarkably, the disparity in the total number of AGTHs generated between the highest and lowest performing talkers reaches 128, underscoring a significant variation in generalization performance across the different talkers.

\begin{table}[!tp]
    \centering
    \caption{Details of talkers employed for generation. }
    \vspace{-0.3cm}
    \resizebox{1\linewidth}{!}{\begin{tabular}{c|c|c|c|c|c}
    \toprule
         Type & Label &Methods & Year & Head Motion & Output Resolution\\ \hline
           \multirow{6}{*}{Image-based} &MI &MakeIttalk \cite{makelttalk} &2020 &\checkmark &256×256 \\  
          &AH&Auido2Head \cite{audio2head}&2021 &\checkmark &256×256 \\
        &ST&Sadtalker \cite{sadtalker}&2023 &\checkmark &512×512 \\
         &DT&Dreamtalk \cite{dreamtalk}&2023 &\checkmark &256×256  \\
         &ET&EAT \cite{eat}&2023 &\checkmark &256×256 \\ 
          &EM&EchoMimic \cite{chen2024echomimic}&2024 &\checkmark &512×512 \\ 
           \hdashline
          \multirow{8}{*}{Video-based}&WL&Wav2Lip \cite{wav2lip}&2020 &\ding{53} &1,024×1,024 \\
         &VR&Video-Retalking \cite{videoretalking}&2022 &\ding{53} &1,024×1,024 \\
         &SH&StyleHeat \cite{yin2022styleheat}&2022 &\ding{53} &1,024×1,024 \\ 
         &DN&DINet \cite{dinet}&2023 &\ding{53} &1,024×1,024 \\
          &IL&IP-LAP \cite{iplap}&2023 &\ding{53} &1,024×1,024 \\
        &TL&TalkLip \cite{wang2023seeing}&2023 &\ding{53} &1,024×1,024 \\ 
         &MT&MuseTalk \cite{musetalk}&2024 &\ding{53} &1,024×1,024 \\

          &EG&EmoGen \cite{yang2024emogen}&2024 &\ding{53} &1,024×1,024 \\ 
    \bottomrule
    \end{tabular}}
    \label{tab:talker}
    \vspace{-0.3cm}
\end{table}

To investigate the quality of AGTHs and their distortions in detail, we recruit 13 male and 12 female participants to conduct subjective quality assessments of all AGTHs in the THQA-10K dataset. The subjective evaluation is carried out in a well-controlled laboratory in accordance with the guidelines outlined in ITU-R BT.500-13 \cite{bt2002methodology}. Participants can view AGTHs on an iMac monitor with a resolution of 4,096 × 2,304. Given that the AGTHs include audio components, a wired headset is utilized to ensure low-latency and high-quality audio transmission, while also minimizing potential interference between participants due to the audio output. The AGTHs are organized into 100 phases, with each phase comprising at most 120 AGTHs. To mitigate visual fatigue and discomfort associated with prolonged viewing, all participants are required to take a 15-minute break after completing each phase. Furthermore, each participant is limited to a maximum of 6 assessment phases per day. 

    
\subsection{Data Processing}
At the end of the subjective experiment, we receive a total of 261,425 = 25 × 10,457 subjective ratings. Each rating can be described as $\{s_{ij}, D_{ij}\}$, where $s_{ij}$ and $D_{ij}$ are the quality rating and labeled distortion of the $j$-th AGTH by the $i$-th subject. In particular, $D_{ij}$ is a ten-dimensional 0-1 distortion vector, with each dimension denoting a corresponding distortion type. According to existing works \cite{3dgcqa,reliqa,dhhqa,ddhqa,h3d,zhou2021omnidirectional,zhou2022pyramid,zhou2023quality,min2024perceptual,zhou2023perceptual,zhang2024quality,zhou2025q,zhang2025mm}, $s_{ij}$ is processed as z-scores according to the following equation:
\begin{equation}
z_{ij} = \frac{{{s_{ij}} - \mu _i}}{{\sigma _i}},
\end{equation}
where $\mu_{i}=\frac{1}{N_{i}} \sum_{j=1}^{N_{i}} s_{i j}$, $\sigma_{i}=\sqrt{\frac{1}{N_{i}-1} \sum_{j=1}^{N_{i}}\left(s_{i j}-\mu_{i}\right)}$, and $N_i$ represents the total number of AGTHs evaluated by subject $i$. Following the rejection procedure outlined in \cite{bt2002methodology}, ratings from unreliable subjects are excluded. The remaining z-scores $z_{ij}$ are linearly rescaled to the range [0, 5]. Finally, the mean opinion scores (MOSs) for the $j$-th AGTH are computed by averaging the rescaled z-scores. For the distortion vector $D_{ij}$, the summation operation is employed to count the distortion type of the $j$-th AGTH:
\begin{equation}
D_j=\sum_{i=1}^{N_j}D_{ij},
\end{equation}
where $N_j$ denotes the number of subjects who classified distortions of the $j$-th AGTH. A threshold vector $T_j$, defined as ${N_j}/2$ across all ten dimensions, is utilized to derive the final distortions $\mathcal{D}_j$:
\begin{equation}
\mathcal{D}_j = u(D_j-T_j),
\end{equation}
where $u(\cdot)$ denotes the step function. The whole process can be interpreted as for each category of distortion for each AGTH, more than half of the subjects need to believe that the distortion exists before it is officially acknowledged.

\begin{figure}[!t]
    \centering
    
    \subfloat[]{\includegraphics[width=0.494\linewidth]{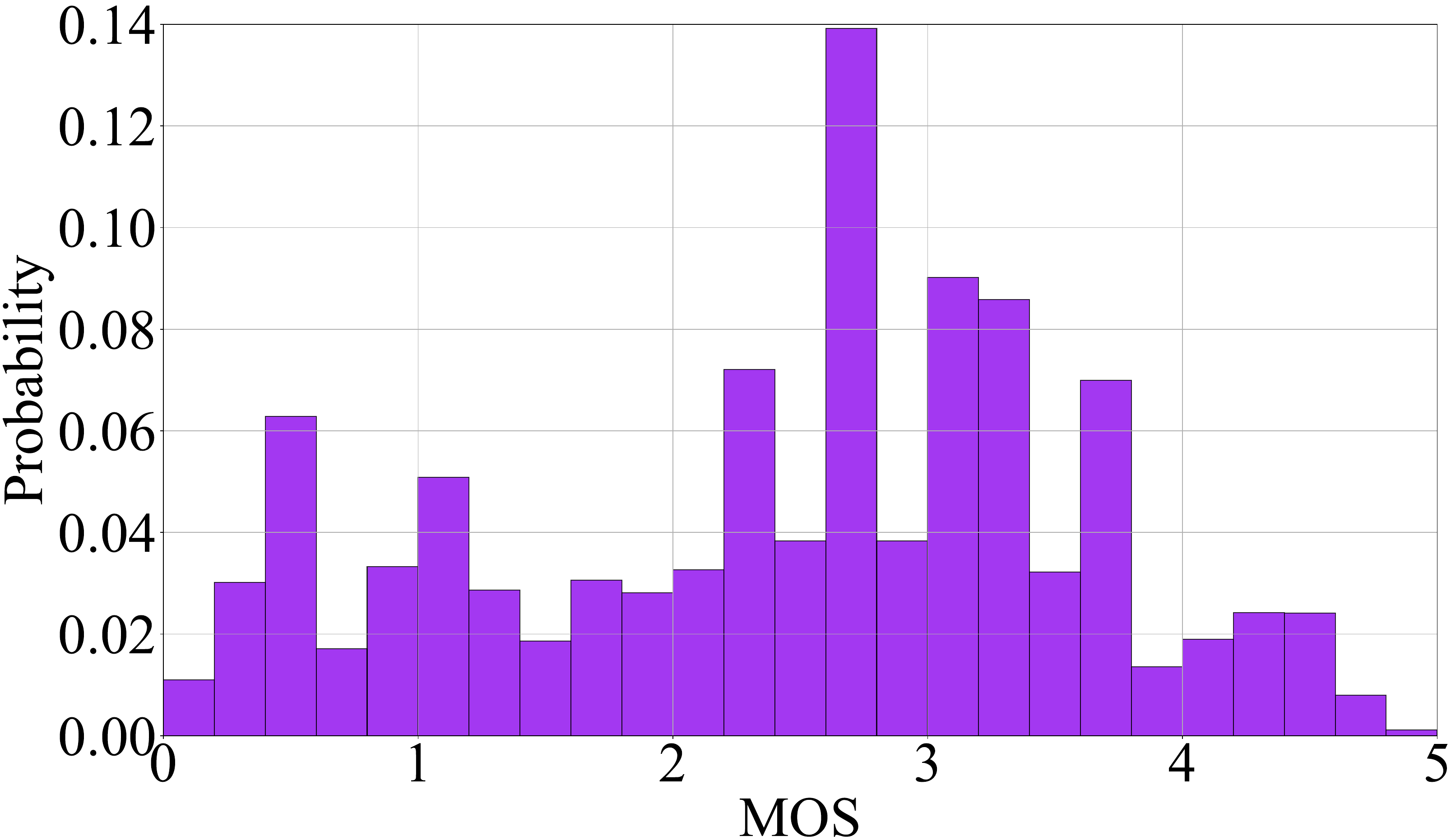}}
    \subfloat[]{\includegraphics[width=0.494\linewidth]{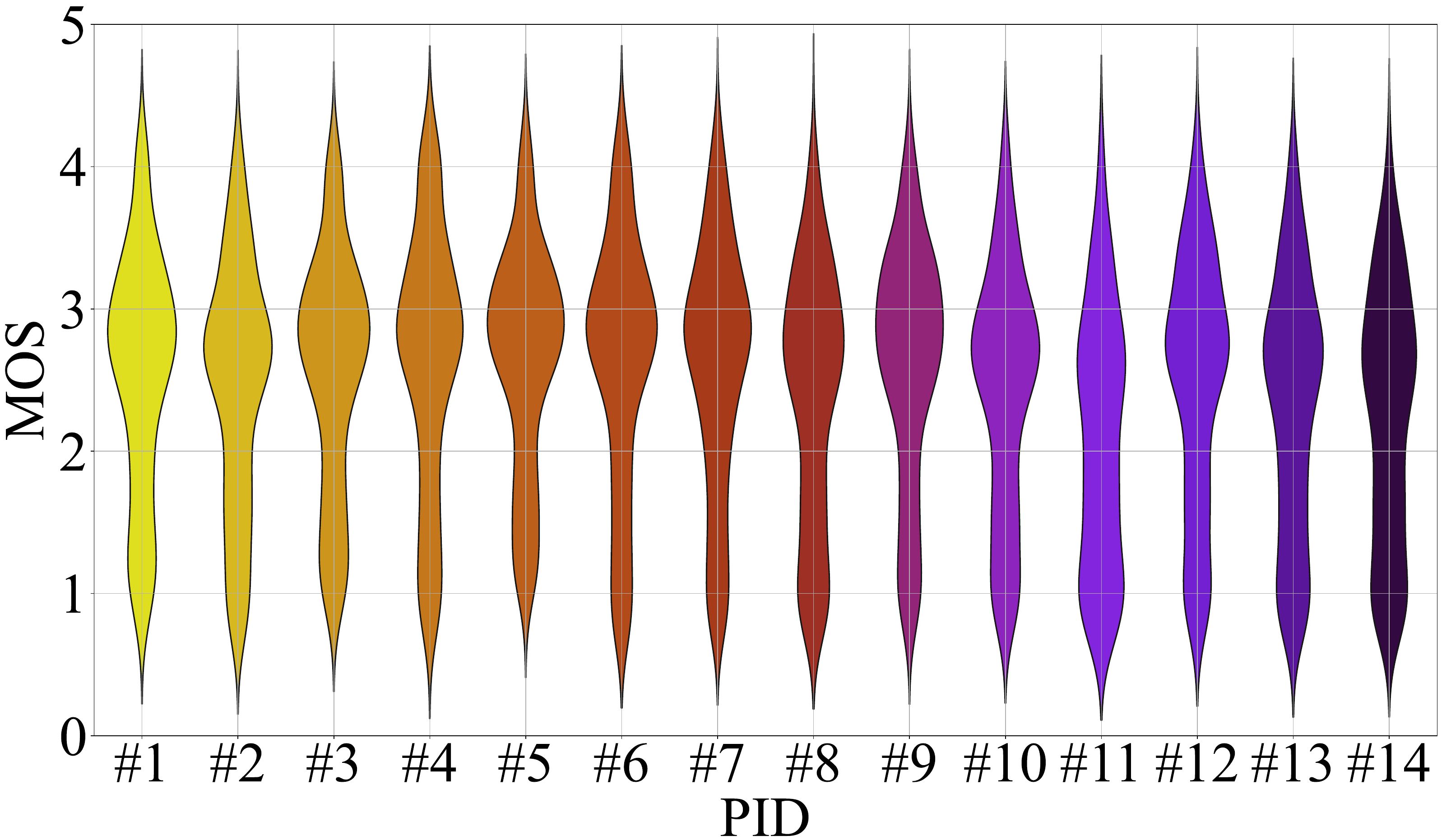}}
    \vspace{-0.1cm}
    \subfloat[]{\includegraphics[width=0.494\linewidth]{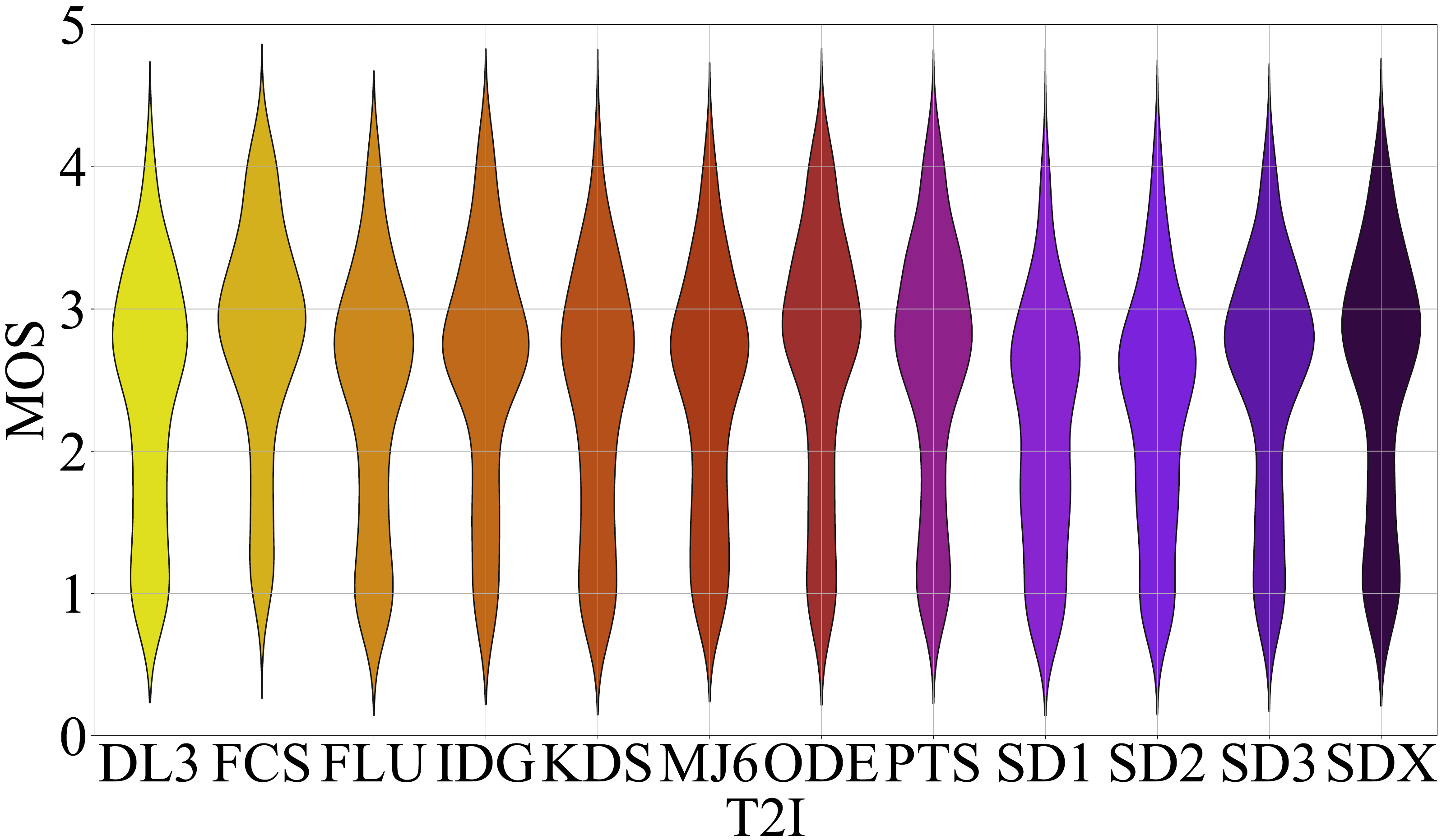}}
    \subfloat[]{\includegraphics[width=0.494\linewidth]{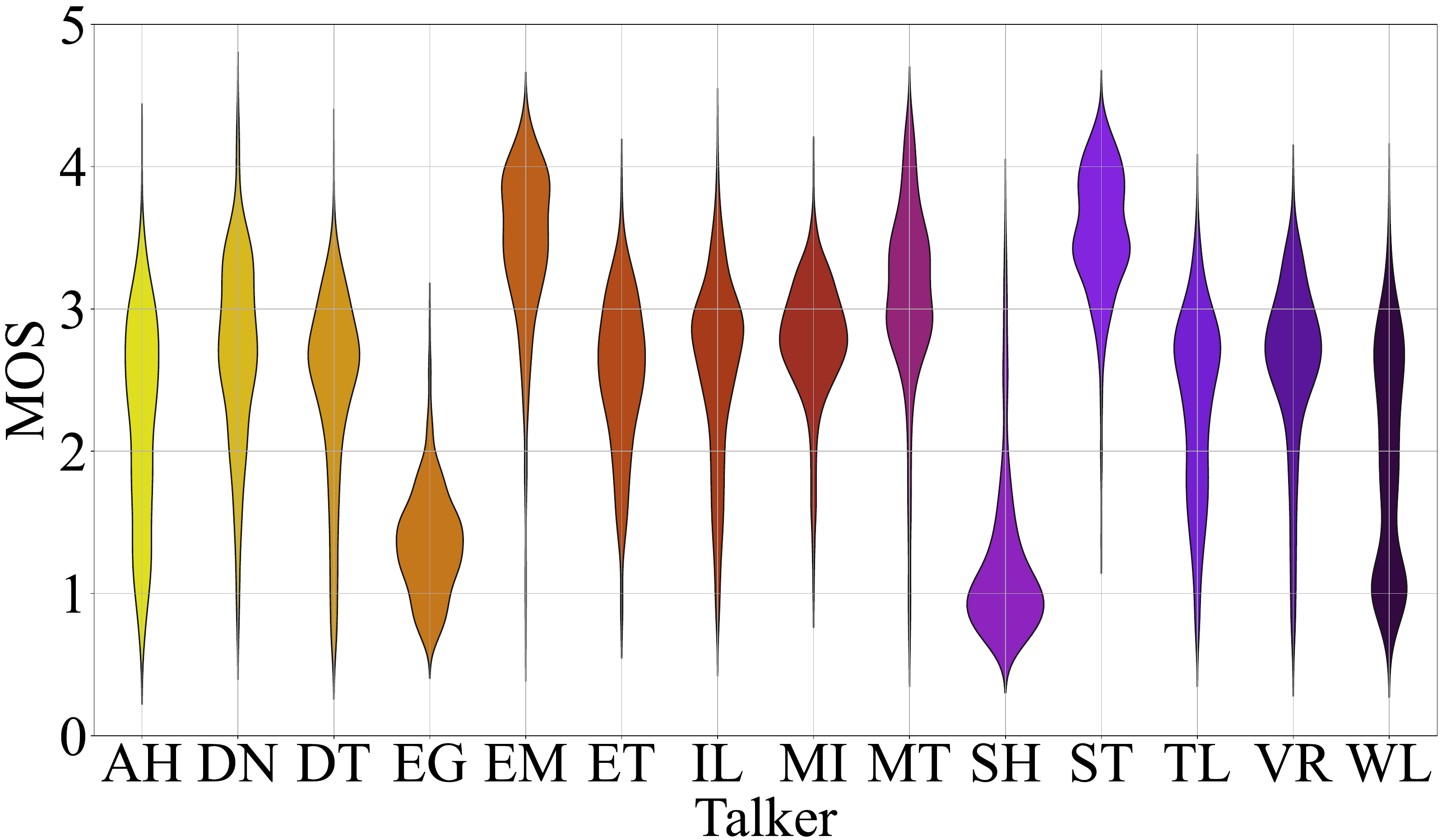}}
    
    \vspace{-0.4cm}
    \caption{Distribution of MOSs.}
    \label{fig:mosviolin}
    \vspace{-0.6cm}
\end{figure}

\begin{figure*}[!t]
    
    \centering
    \includegraphics[width =1\linewidth]{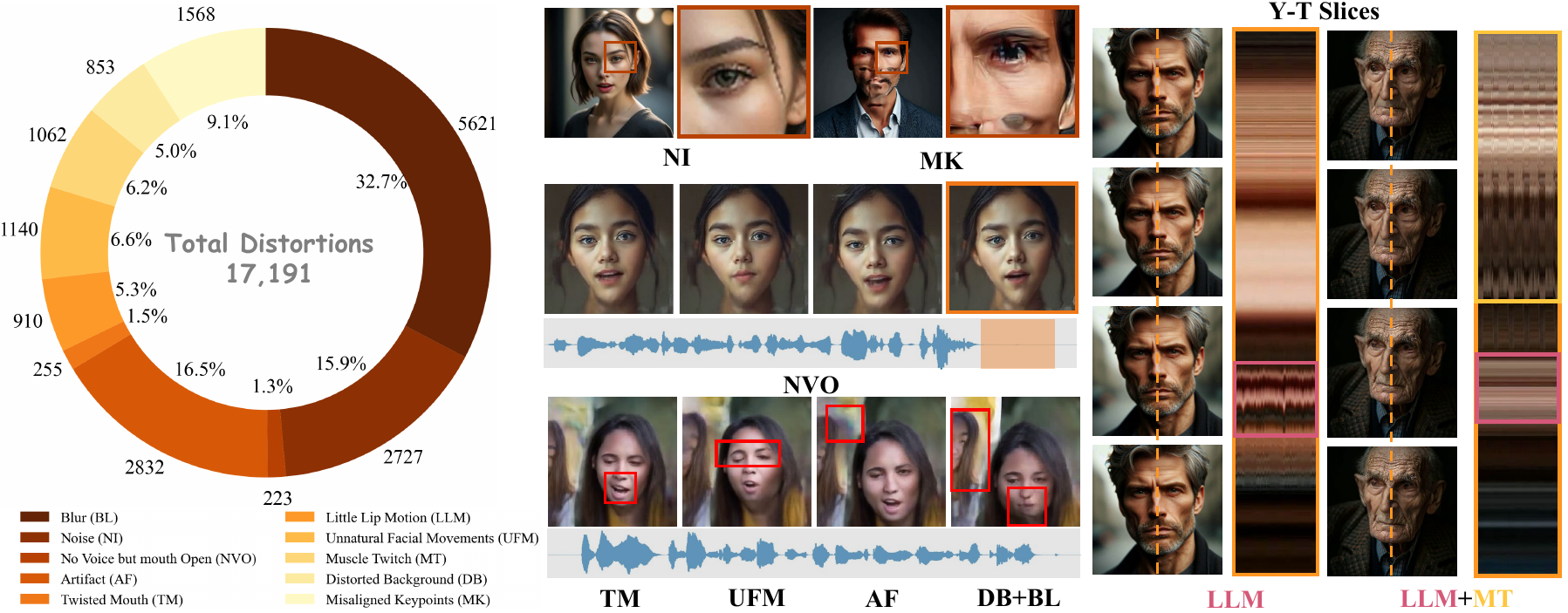}
    \vspace{-0.7cm}
    \caption{Visualization of distortions. The left side shows how often each distortion occurs, while the right side shows typical cases.}
    \label{fig:disvis}
    \vspace{-0.5cm}
\end{figure*}

\begin{figure}[!t]
    \centering
    
    \subfloat[]{\includegraphics[width=0.499\linewidth]{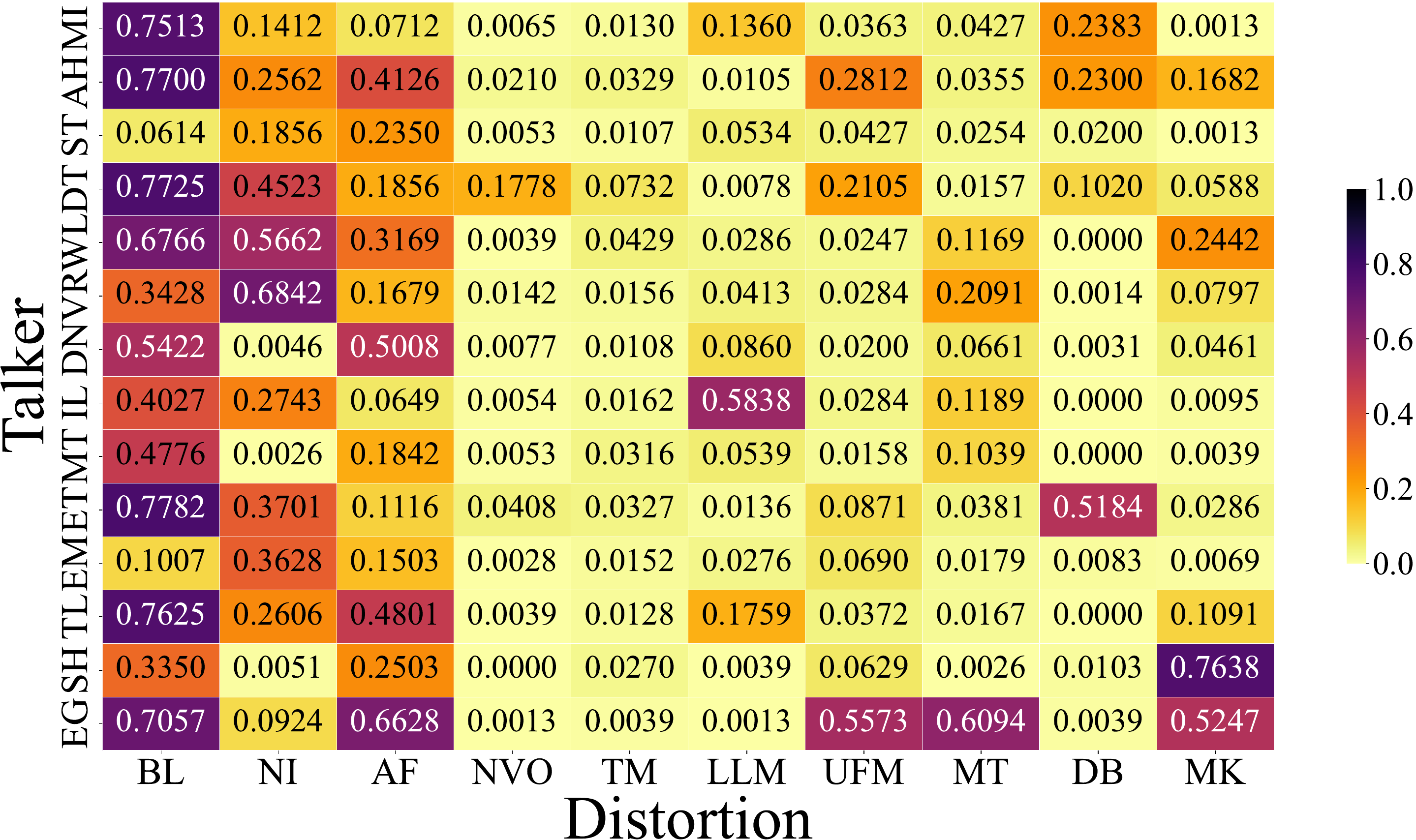}}
    \subfloat[]{\includegraphics[width=0.499\linewidth]{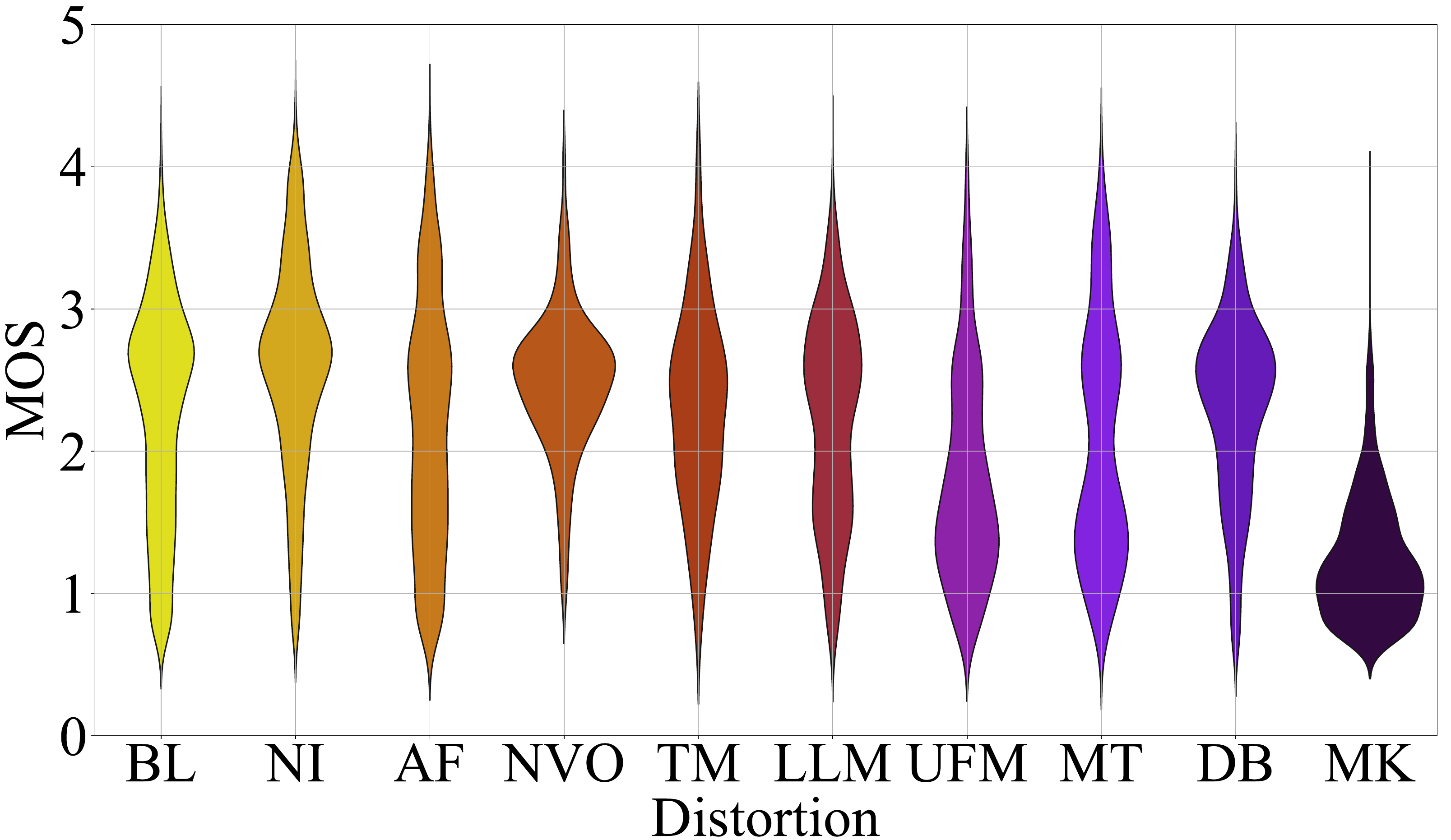}}
    
    \vspace{-0.3cm}
    \caption{Statistical analysis of different distortions.}
    \label{fig:disana}
    \vspace{-0.6cm}
\end{figure}

    

\subsection{Mean Opinion Score Analysis}
\label{sub:mos}
Based on the results of the subjective experiments, we conduct a comprehensive analysis of mean opinion scores (MOSs) and distortion of AGTHs in order to evaluate the various types of talkers. Initially, the overall MOS distribution is plotted as shown in Fig.~\ref{fig:mosviolin}(a). To further analyze the effects of various possible factors on the MOSs, the violin plots shown in Fig.~\ref{fig:mosviolin}(b-d) provide a more intuitive picture of the relationship between different PIDs, T2Is, talkers and MOSs. By observing Fig.~\ref{fig:mosviolin}, we can draw some valuable conclusions: 1) The majority of AGTHs received quality scores centered around 3, indicating that some talkers are capable of meeting user expectations regarding audiovisual quality. However, it is noteworthy that only a limited number of AGTHs achieved MOSs exceeding 4 points, while a significant proportion of AGTHs received low-quality scores in the range of 0 to 2 points. This suggests considerable potential for improving the audiovisual quality produced by talkers; 2) AGTHs generated from different PIDs and T2Is exhibit a similar distribution of MOSs. This observation indicates that the PIDs and T2Is utilized in constructing the THQA-10K dataset are both representative and universal, thereby facilitating generalization to other PIDs and T2Is. Additionally, it implies that variations in PIDs and T2Is are not the primary determinants affecting the quality of AGTHs; 3) There are significant disparities in the quality distribution of AGTHs produced by different talkers. Notably, two types of talkers, EM and ST, generate the highest quality AGTHs, whereas two other types, EG and SH, yield lower quality outputs.
    

\begin{figure*}[!t]
    
    \centering
    \includegraphics[width =1\linewidth]{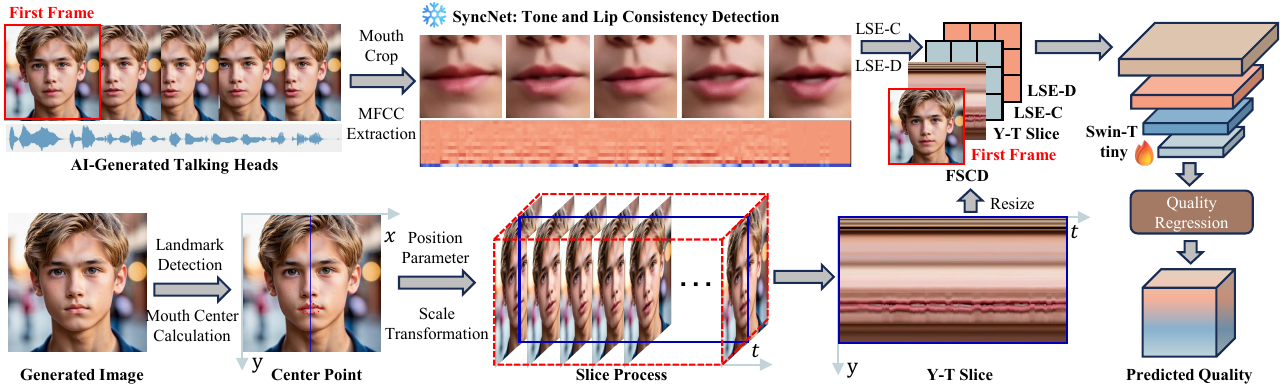}
    \vspace{-0.6cm}
    \caption{The proposed framework for FSCD. The method mainly consists of four compartmentalized modules: the Y-T slice process, tone-lip consistency detection, backbone feature extraction, and quality regression.}
    \label{fig:fscd}
    \vspace{-0.4cm}
\end{figure*}

\subsection{Distortion Visualization \& Analysis}
An examination of the AGTHs in the THQA-10K dataset reveals that the 9 distortion types previously identified in the THQA dataset by Zhou $et$ $al.$ \cite{thqa} remain prevalent. Additionally, new distortions have emerged, predominantly characterized by misalignment of facial keypoints. To illustrate the impact of each distortion type and quantify their occurrence, we select representative samples for visualization and cataloged the distortions identified in the subjective assessments. The results are depicted in Fig.~\ref{fig:disvis}. From this analysis, several key insights can be derived: 1) A total of 17,191 distortions are identified across 10,457 AGTHs, indicating that distortions are a common issue within existing AGTHs. Notably, individual AGTHs often exhibit a combination of multiple distortion types; 2) Among the various distortion types, blur (BL), noise (NI) and artifacts (AF) remain the predominant issues, suggesting that current talkers still face limitations in these areas; 3) The use of Y-T slices to represent the two distortion types, little lip motion (LLM) and muscle twitch (MT), as demonstrated by Zhou $et$ $al.$, offers a more pronounced indication of the visual effects associated with these distortions. Specifically, LLM manifests as parallel lines or minor fluctuations in the mouth texture within the Y-T slice, while MT is characterized by periodic repetitive textures. The newly identified misaligned keypoints (MK) distortion results in a noticeable displacement of facial features, leading to severe distortion.

To further investigate the frequency of various distortion types across different talkers, we plot a heatmap to visually illustrate the relationship between talkers and distortions. The data presented in Fig.~\ref{fig:disana}(a) reveal another two key findings: 1) Specific distortions tend to be concentrated among one or a few talkers. For instance, LLM predominantly occurs with the IL talker, while distortion from distorted background (DB) is mainly observed with the ET talker, and MK is frequently associated with the SH and EG talkers; 2) Among all talkers included in the comparison, ST and EM exhibit the least frequency of distortion, highlighting their superiority and explaining why these two methods achieve the highest MOSs. Finally, in conjunction with the MOS analysis method of Sec.~\ref{sub:mos}, the effect of each type of distortion on the MOSs is plotted as Fig.~\ref{fig:disana}(b). The results indicate that different distortion types have varying effects on human audiovisual perception. Notably, the three most common distortions do not significantly detract from the overall quality of AGTHs, and in some cases, they still yield high MOSs. Conversely, MK has the most pronounced negative impact on AGTH quality, primarily due to the intolerable displacement of facial features, resulting in significant distortion.

\section{Proposed Method}
\subsection{First Frame \& Slice Process}
Considering that each AGTH video is generated by a stationary talker, it is reasonable to assert that the video quality remains relatively stable throughout the duration of the AGTH. Consequently, the initial frame of an AGTH provides a rich set of spatial features suitable for quality assessment. In terms of temporal feature selection, this paper proposes a slice-based temporal feature extraction scheme, as illustrated in Fig.~\ref{fig:fscd}. Firstly, we advocate for the use of Y-T slices over X-T slices, as Y-T slices capture more comprehensive facial information due to the inherent symmetry of the face. Additionally, to ensure that the tangent line intersects the mouth, which is a critical aspect of AGTHs, we utilize the image generated by the T2I model as a reference and apply landmark detection to identify the key points $p_k$ of the mouth within the generated image. Subsequently, we compute the coordinates of the mouth's centroid $C_o$:
\begin{equation}
{C_o} = \frac{1}{K}\sum\nolimits_{k = 1}^K {{p_k}},
\end{equation}
where $K$ denotes the total number of key points on the face. However, this derived center point cannot be directly employed to guide the slicing of corresponding AGTHs, as the videos produced by different talkers exhibit varying resolutions $R_v$. Therefore, a scale transformation is applied to determine the location of the mouth centroid in the AGTH videos:
\begin{equation}
{C_v} = \frac{{{R_v}}}{{{R_o}}}{C_o},
\end{equation}
where $R_o$ is the resolution of the generated portrait and $C_v$ denotes the center of the human face in AGTHs. Following this, the entire AGTH video can be projected as a Y-T slice within the three-dimensional space defined by XY-T coordinates. Ultimately, the Y-T slice is resized to match the resolution of the first frame, facilitating subsequent processing. This entire slicing process underscores that the Y-T slice effectively captures temporal features over the video duration, in contrast to conventional frame extraction methods.

\subsection{Tone-lip Consistency Detection}
AGTH is a media that exhibits a high sensitivity to audio-visual synchronization, particularly in relation to the synchronization between lip movements and speech. To address this sensitivity, we initially perform a cropping of the mouth region in the AGTH and extracted Mel-frequency cepstral coefficients (MFCCs) from the accompanying audio. Subsequently, we employ the classical SyncNet \cite{chung2017out} to assess audio-lip consistency, yielding two key outputs: lip sync error confidence (LSE-C) and distance (LSE-D). In contrast to existing methodologies that integrate these two features directly into the final quality regression layer, this study proposes an innovative approach whereby the two LSE features are expanded into a tensor of dimensions corresponding to those of the first frame. This tensor is then utilized as two additional images within the first frame, enabling the backbone network to learn the relative significance of tonal lip consistency autonomously, rather than relying on predetermined weights based on prior knowledge.
\subsection{Backbone \& Quality Regression}
The obtained \underline{F}irst frame, Y-T \underline{S}lice, LSE-\underline{C}, and LSE-\underline{D} can be collectively viewed as a new image called FSCD. Given the excellent performance achieved by the swin-transformer (Swin-T) \cite{liu2021swin,zhou2025ct,zhang2023simple,zhang2023eep} in several computer vision tasks, Swin-T is used to extract quality features from FSCDs. During the training phase, the predicted audiovisual quality is compared with the actual MOS using the Mean Squared Error (MSE) as the loss function, facilitating the gradual optimization of the algorithm for enhanced performance:

\begin{equation}
Loss = \frac{1}{n}\sum\limits_{l = 1}^n {{{({{\hat Q}_l} - {Q_l})}^2}} ,
\end{equation}
where $\hat Q_{l}$ and $Q_l$ represent the predicted quality and MOS of $l$th AGTH, and $n$ indicates the size of training batch.

\begin{table*}[!t]
    \centering
    \setlength{\tabcolsep}{5pt}
    \caption{Performance results on the proposed THQA-10K, THQA and THQA-3D databases. Best in {\bf\textcolor{red}{RED}}, second in \bf\textcolor{blue}{BLUE}.}
    \vspace{-0.3cm}
    \resizebox{\linewidth}{!}{\begin{tabular}{c|l|cccc|cccc|cccc}
    \toprule
       \multirow{2}{*}{Type} &\multirow{2}{*}{Models}   &  \multicolumn{4}{c|}{THQA-10K} &  \multicolumn{4}{c|}{THQA} &\multicolumn{4}{c}{THQA-3D}\\ \cline{3-14}
       &   & SRCC$\uparrow$ & PLCC$\uparrow$ & KRCC$\uparrow$ & RMSE$\downarrow$ & SRCC$\uparrow$ & PLCC$\uparrow$ & KRCC$\uparrow$ & RMSE$\downarrow$ &SRCC$\uparrow$ & PLCC$\uparrow$ & KRCC$\uparrow$ & RMSE$\downarrow$\\ \hline
       \multirow{4}{*}{IQA} 
        & BRISQUE \cite{brisque} & 0.4271& 0.4451&0.2993 &1.0262 & 0.4856 & 0.5970& 0.3454&0.8227 & 0.6749& 0.7453&0.5060 &0.5717\\
       & NIQE \cite{niqe} & 0.0089&0.0436 &0.0051 & 1.1492&0.0535	&0.1643	&0.0402	&0.9811 & 0.2243&0.4741 & 0.1232&0.7707\\
       & CPBD \cite{cpbd} &0.0553 &0.0686 &0.0371 & 1.1476&	0.0575	&0.0876	&0.0376	&0.9908 &0.2145 &0.3136 &0.1432 &0.8273\\
       & IL-NIQE \cite{ilniqe} & 0.0490& 0.0634& 0.0286& 1.1480& 0.0537	&0.2160	&0.0276	&0.9712 & 0.2293&0.4871 & 0.1537&0.7600\\ \hdashline
       \multirow{2}{*}{Sync} 
       & LSE-C \cite{chung2017out}  & 0.0706& 0.1634& 0.0468& 1.1349&	0.0056	&0.2109	&0.0048	&0.9723 &0.1728 &0.2297 &0.1355 &0.8499\\
       & LSE-D \cite{chung2017out} & 0.0580& 0.1123& 0.0385& 1.1431&	0.1366	&0.2336	&0.0855	&0.9671 & 0.0079 &0.1054 &0.0008 &0.8684\\  \hdashline
       \multirow{10}{*}{VQA} & VIIDEO \cite{viideo} &0.1354 &0.1782 & 0.0901& 1.1319& 0.1777 &0.1891 & 0.1354& 0.9595& 0.1056& 0.2308& 0.0721&  0.8387\\
       & TLVQM \cite{tlvqm} &0.4377 & 0.4679& 0.3070&1.0130 & 0.0254 &0.0355& 0.0209&1.0853& 0.1887& 0.3112 &0.1272 &0.8240\\
       & VIDEVAL \cite{videval} & 0.3869&0.4147 &0.2706 & 1.0431& 0.0317 & 0.0358& 0.0231&1.1916&0.2252 & 0.3544& 0.1556& 0.8118\\
       & V-BLIINDS \cite{vblinds} & 0.4740& 0.4977& 0.3334&0.9941 & 0.4949 & 0.6403&0.3533&0.7976 &0.5298& 0.6412& 0.3907&0.6674\\
       & RAPIQUE \cite{rapique} & 0.3576& 0.3846& 0.2490& 1.0579& 0.1789 & 0.1908& 0.1277&1.0162 & 0.3748& 0.4680& 0.2660&0.7643\\
       & SimpVQA \cite{simpvqa} & \bf\textcolor{blue}{0.7775}& \bf\textcolor{blue}{0.8039}& \bf\textcolor{blue}{0.5931}& \bf\textcolor{blue}{0.6832}& 0.6800 & 0.7592 & 0.5052 & 0.6361&0.6321   & 0.7258&0.4717 &0.5983\\
       & VSFA \cite{vsfa} & 0.7537& 0.7754& 0.5726&0.7343&\bf\textcolor{blue}{0.7601} & \bf\textcolor{blue}{0.8106}& \bf\textcolor{blue}{0.5830}&\bf\textcolor{blue}{0.5966} & 0.7463&0.7811  &0.5596 &0.5726\\
       & FAST-VQA \cite{fastvqa} & 0.7351& 0.7542& 0.5519& 0.8026& 0.6389 & 0.7441& 0.4677&0.6983 & 0.7778&0.7984 &0.5964 &\bf\textcolor{blue}{0.5503}\\
      & BVQA \cite{bvqa} &0.6335 & 0.7405& 0.4522& 0.7634& 0.7287 & 0.7985&0.5549 &0.6094 & \bf\textcolor{blue}{0.7871}&\bf\textcolor{blue}{0.8298} &\bf\textcolor{blue}{0.6081} &0.5983\\ \hdashline
          \multicolumn{2}{c|}{\textbf{FSCD (Ours)}}  & \bf\textcolor{red}{0.8066}& \bf\textcolor{red}{0.8322}& \bf\textcolor{red}{0.6228}&\bf\textcolor{red}{0.6333} &\bf\textcolor{red}{0.7812} & \bf\textcolor{red}{0.8409} &\bf\textcolor{red}{0.5951} &\bf\textcolor{red}{0.5055} &\bf\textcolor{red}{0.8235} &\bf\textcolor{red}{0.8505} &\bf\textcolor{red}{0.6463} & \bf\textcolor{red}{0.4577}\\
    \bottomrule
    \end{tabular}}
    \vspace{-0.4cm}
    \label{tab:performance}
\end{table*}

\begin{table*}[!t]
         \caption{Ablation study results on databases, where `\textit{w/o}' stands for `without'. Best in {\bf\textcolor{red}{RED}}, second in {\bf\textcolor{blue}{BLUE}}. }
         \vspace{-0.3cm}
    \resizebox{1\linewidth}{!}{\begin{tabular}{c|cccc|cccc|cccc}
    \toprule
       \multirow{2}{*}{Dimension}  &\multicolumn{4}{c|}{THQA-10K} &\multicolumn{4}{c|}{THQA} &\multicolumn{4}{c}{THQA-3D}\\ \cline{2-13}
        & SRCC$\uparrow$ & PLCC$\uparrow$ & KRCC$\uparrow$ & RMSE$\downarrow$ & SRCC$\uparrow$ & PLCC$\uparrow$ & KRCC$\uparrow$ & RMSE$\downarrow$ & SRCC$\uparrow$ & PLCC$\uparrow$ & KRCC$\uparrow$ & RMSE$\downarrow$\\ \hline
        \textit{w/o} $F$  &0.6510 &0.6892 & 0.4726 &0.8497 & 0.6830 &0.7506 &0.5035 &0.5765 &0.7368 &0.7844 & 0.5451 & 0.5468   \\
         \textit{w/o} $S$ & 0.7330& 0.7672& 0.5509& 0.7329&0.7205 & 0.7968 & 0.5325 & 0.5272 & 0.7506 & 0.7730 & 0.5655 & 0.5593\\
        \textit{w/o} $C$  & \bf\textcolor{blue}{0.7927}& \bf\textcolor{blue}{0.8169}& \bf\textcolor{blue}{0.6049}& \bf\textcolor{blue}{0.6764}&\bf\textcolor{blue}{0.7660} & \bf\textcolor{blue}{0.8174}&\bf\textcolor{blue}{0.5860} &\bf\textcolor{blue}{0.5086} & \bf\textcolor{blue}{0.8019} &\bf\textcolor{blue}{0.8429} &\bf\textcolor{blue}{0.6160} &\bf\textcolor{blue}{0.4658}\\
        \textit{w/o} $D$  & 0.7610& 0.7879& 0.5744& 0.7221&0.7462 &0.8110 &0.5579 &0.5105 &0.7915 & 0.8359 & 0.6037 & 0.4710\\ \hline
        FSCD  & \bf\textcolor{red}{0.8066}& \bf\textcolor{red}{0.8322}  & \bf\textcolor{red}{0.6228} & \bf\textcolor{red}{0.6333} &\bf\textcolor{red}{0.7812} & \bf\textcolor{red}{0.8409} & \bf\textcolor{red}{0.5951} &\bf\textcolor{red}{0.5055} &\bf\textcolor{red}{0.8235} &\bf\textcolor{red}{0.8505} &\bf\textcolor{red}{0.6463} & \bf\textcolor{red}{0.4577}\\ 
       
    \bottomrule

    \end{tabular}}
    \vspace{-0.5cm}
 \label{tab:abl}
    \end{table*}
    
\section{Experiments}
\subsection{Experiment Details \& Criteria}

To validate the effectiveness of the proposed method, we select 15 quality assessment algorithms applicable to AGTHs for comparison. This selection includes 4 classical image quality assessment (IQA) methods, 2 methods for audio and lip consistency, and 9 video quality assessment (VQA) methods. Among these, RAPIQUE, SimpVQA, VSFA, FAST-VQA, and BVQA are deep learning-based methods, while the remaining methods rely on manually extracted features. All selected methods are tested on THQA-10K, THQA \cite{thqa} and THQA-3D \cite{thqa3d} datasets, utilizing a five-fold cross-validation scheme. The average test results from the five folds are recorded as the performance of each method. Notably, the five-fold data partitioning ensures that there is no content overlap, and all algorithms employed are derived from the source code provided by their authors.

In terms of evaluation criteria, we adopt four commonly used metrics for assessing the performance of objective multimedia quality assessment algorithms: Spearman Rank Correlation Coefficient (SRCC), Kendall’s Rank Correlation Coefficient (KRCC), Pearson Linear Correlation Coefficient (PLCC), and Root Mean Squared Error (RMSE).

\subsection{Performance Analysis}
The performance of the proposed FSCD method, along with other competing methods, on three datasets is presented in Table~\ref{tab:performance}. An analysis of this table yields several key conclusions: 1) FSCD demonstrates optimal performance on all datasets, surpassing the next best algorithm by +2\% in SRCC at least. This result strongly supports the effectiveness of the proposed FSCD method for assessing AGTH quality; 2) The methods that achieve suboptimal performance differ among three datasets. In contrast, FSCD consistently achieves optimal performance across all datasets, highlighting the robustness and generalizability of the proposed method; 3) Existing IQA methods, single audio-lip consistency detection techniques, and VQA methods are constrained in their ability to evaluate AGTH quality. This limitation primarily arises from the inability of these methods to fully leverage the spatio-temporal features and multimodal information inherent in AGTHs. Moreover, compared to the extraction of temporal features through the Y-T slice, conventional approaches often struggle to adequately account for the relationships among frames.


\subsection{Ablation Experiments}
To further evaluate the effectiveness of each component within the FSCD framework, ablation experiments are conducted, with results summarized in Table~\ref{tab:abl}, from which following observations can be drawn: 1) Each component of FSCD contributes positively to the overall performance. This enhancement can be attributed to the fact that the four components address different quality aspects, allowing for a synergistic effect; 2) Regarding the importance of individual components, the quality characteristics provided by the first frame rank highest, followed by the Y-T Slice, LSE-D, and LSE-C. This suggests that spatial features are paramount in influencing the quality of AGTHs, with temporal features and coherence features following in importance; 3) A comparative analysis of the performance results presented in Tables ~\ref{tab:performance} and~\ref{tab:abl} reveals a noteworthy finding: even when utilizing only the Y-T Slice and the tone and lip coherence features, without the first frame, competitive performance is still achievable on three datasets. This highlights the validity and significance of the Y-T slice as employed by FSCD in the assessment of AGTH quality.
\section{Conclusion}
As digital human technology continues to advance, various speech-driven methods, commonly referred as ``Talkers," have emerged to enhance the efficiency of digital human face design. To thoroughly investigate the generalization performance and generation quality of different talkers, this study introduces the THQA-10K dataset, comprising a total of 10,457 AI-Generated Talking Head (AGTH) videos. Specifically, 12 Text-to-Image (T2I) methods are employed to generate character portraits, while 14 advanced talkers are utilized to produce AGTHs. Through comprehensive data analysis and subjective experiments conducted on the THQA-10K dataset, we validate both the comprehensiveness and generalizability of the dataset. Additionally, we assess the generalization performance of existing talkers and identify potential quality issues and their distributions. Finally, we propose an objective quality assessment method named FSCD, leveraging the first frame, Y-T slice, and tone-lip consistency. Experimental results substantiate the effectiveness and robustness of FSCD, which is anticipated to inform the ongoing development of talkers.

\small{\paragraph{Acknowledgements.}
This work was supported in part by  the Major Key Project of PCL (PCL2023A10-2), National Natural Science Foundation of China (623B2073, 62101326, 62225112, 62271312, 62132006) and STCSM (22DZ2229005).
}
\newpage
{
    \small
    \bibliographystyle{ieeenat_fullname}
    \bibliography{main}
}

\end{document}